\documentclass[letterpaper, 10 pt, conference]{IEEEtran}  

\IEEEoverridecommandlockouts                              

\usepackage{graphicx} 
\usepackage{amssymb, amsmath, amssymb, amsfonts, booktabs, multirow}  
\usepackage{hyperref}
\usepackage{booktabs} 
\usepackage{tabularx} 
\usepackage{multirow} 
\usepackage{caption}
\usepackage{placeins}

\newcolumntype{C}{>{\centering\arraybackslash}X} 

\title{\LARGE \bf
Learning Smooth and Robust Space Robotic Manipulation of Dynamic Target via Inter-frame Correlation
}

\author{Siyi Lang$^{1}$, Hongyi Gao$^{1}$, Yingxin Zhang$^{1}$, Zihao Liu$^{1}$, Hanlin Dong$^{2}$,  \\
Zhaoke Ning$^{3}$, Zhiqiang Ma$^{1}$, and Panfeng Huang$^{1}$
\thanks{*This study was supported by the National Natural Science Foundation of China under Grants 62422316, 62373305, and 62403387, the Fundamental and Interdisciplinary Disciplines Breakthrough Plan of the Ministry of Education of China JYB2025XDXM207. (Corresponding authors: Zhiqiang Ma).}
\thanks{$^{1}$ S. Lang, H. Gao, Y. Zhang, Z. Liu, Z. Ma and P. Huang are with Research Center for Intelligent Robotics, School of Astronautics, Northwestern Polytechnical University, Xi'an 710072, China
.}%
\thanks{$^{2}$H. Dong is with School of Automation, Northwestern Polytechnical University, Xi'an 710072, China
.}%
\thanks{$^{3}$Z. Ning is with School of Aeronautics and Astronautics, Sichuan University, Chengdu 610065, China
.}%
}

\begin{document}

\maketitle
\thispagestyle{empty}
\pagestyle{empty}

\begin{abstract}
On-orbit servicing represents a critical frontier in future aerospace engineering, with the manipulation of dynamic non-cooperative targets serving as a key technology. 
In microgravity environments, objects are typically free-floating, lacking the support and frictional constraints found on Earth, which significantly increases the complexity of tasks involving space robotic manipulation. 
Conventional planning and control-based methods are primarily limited to known, static scenarios and lack real-time responsiveness. 
To achieve precise robotic manipulation of dynamic targets in unknown unstructured space environments, this letter proposes a data-driven space robotic manipulation approach that integrates historical temporal information and inter-frame correlation mechanisms. 
By exploiting the temporal correlation between historical and current frames, the system can effectively capture motion features within the scene, thereby producing stable and smooth manipulation trajectories for dynamic targets. 
To validate the effectiveness of the proposed method, we developed a ground-based experimental platform consisting of a PIPER\_X robotic arm and a dual-axis linear stage, which accurately simulates microgravity free-floating motion in a 2D plane.
\end{abstract}

\section{INTRODUCTION}

Conventional methods for space robotic manipulation primarily rely on planning and control strategies based on dynamic models and human experience~\cite {c1,c2}. These model-based methods require pre-designed task workflows and lack autonomy, making them unsuitable for high-dynamic scenarios that need real-time responses as humans do~\cite{c1,c2}. A possible solution is to use data-driven intelligent robotic manipulation methods~\cite{c33}. By training on a large number of task trajectories, high-precision, reliable, and stable policies can be built~\cite{c28}. 
This approach does not rely on high-accuracy dynamic models or precise target information. It offers certain generalization and multi-task capabilities, representing a promising path for future space robotic manipulation.

Data-driven methods for intelligent robotic manipulation can be mainly divided into two technical paths: Conditional Variational Autoencoder (CVAE) based and Diffusion Policy (DP) based methods~\cite{c3, c4, c5}. CVAE-based approaches generate predicted trajectories through action encoding and conditional decoding, and the inclusion of latent variables provides these methods with good generalization and helps mitigate covariate shift issues. 
Meanwhile, DP models employ active noise addition and conditional denoising to generate a precise motion sequence with strong robustness and high prediction accuracy. 
Another prominent technical path is Vision-Language-Action (VLA) models~\cite{c6,c7,c8,c9}, which have demonstrated excellent performance in various complex tasks, and these models typically possess a massive number of parameters, making them unsuitable for onboard devices with limited computing power.

Existing methods succeed in various tasks, yet conventional models mentioned above, relying on single-frame observations, struggle to capture scene motion features, limiting future state prediction. In imitation learning, one-to-many mapping from branching trajectories induces a multi-modal action distribution~\cite{c35}.

\begin{figure}[h]
    \centering
    \includegraphics[width=\linewidth]{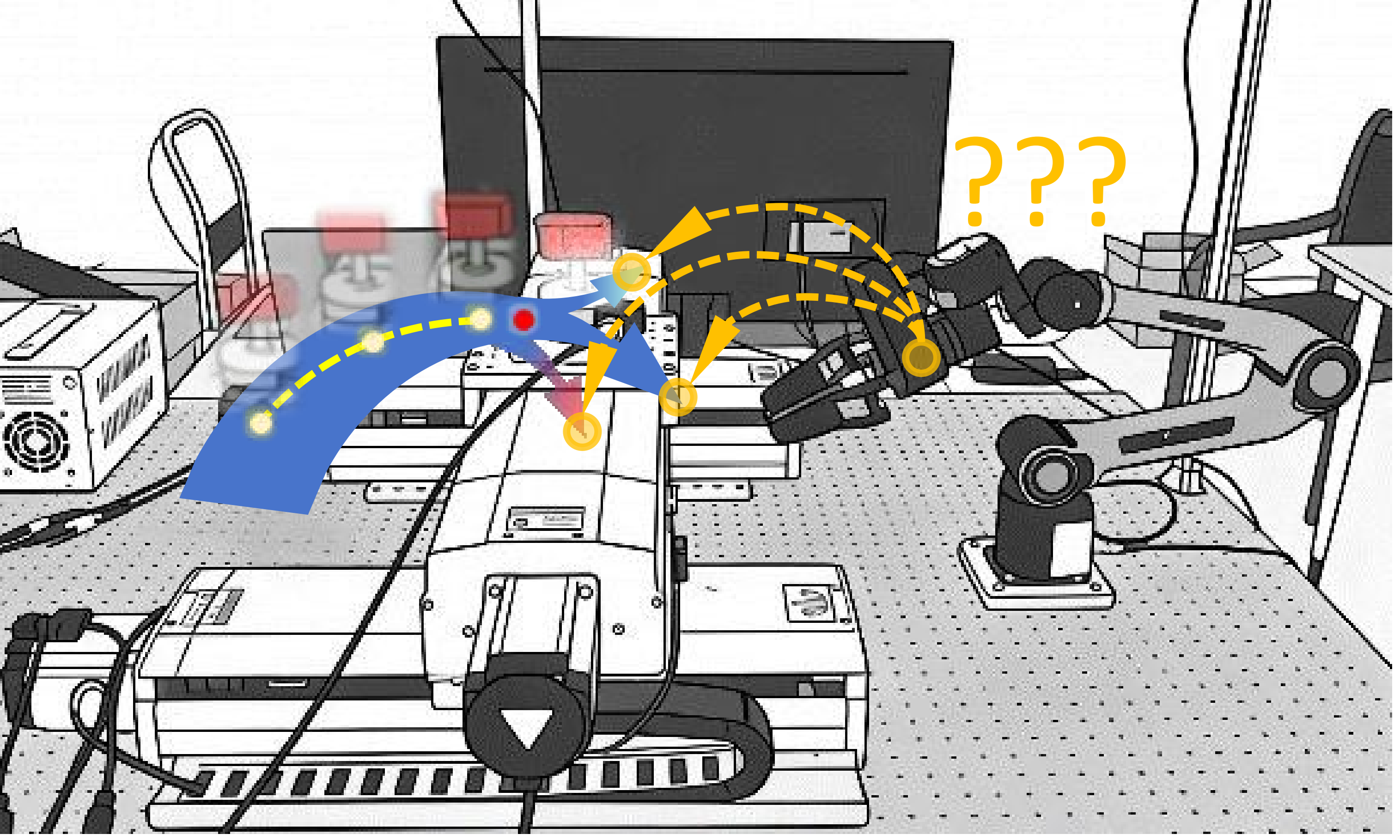}
    \caption{Illustration of trajectory oscillation.}
    \label{fig:TrialsComparison}
\end{figure}

As shown in Fig.~\ref{fig:TrialsComparison}, when a robotic manipulator is grasping a moving target, without temporal consistency, the trained agent may randomly select different grasp points at consecutive time steps, leading to unstable manipulation actions. Connecting the first steps of these inconsistent plans results in severe oscillations and saw-tooth trajectories. Such high-frequency jitter not only reduces the success rate but also compromises the robotic manipulator base's attitude stability through dynamic coupling~\cite{c17}. Given that target dynamics are crucial for successful manipulation, we propose an imitation learning method that explicitly captures inter-frame motion features to enable robust manipulation of dynamic targets.

To address the challenge of learning smooth and robust space robotic manipulation of dynamic targets, we propose a manipulation method based on imitation learning that integrates an Inter-frame Correlation Network. The core innovation of this method is the introduction of an inter-frame correlation mechanism to capture spatio-temporal correlation information within continuous visual streams. By constructing a cost volume~\cite{c10,c11}, the model transforms pixel-level displacement cues between two consecutive frames into high-dimensional action tokens for a Transformer-based action generation network~\cite{c30,c32}. This allows the model to learn the representation of the target's motion trends. This design enables the system to predict the trajectory evolution of the target, thereby achieving precise and stable proactive interception manipulation of dynamic targets.

\section{Related works}

\subsection{Robotic Manipulation of Dynamic targets}

Unlike explicit planning, learning-based systems map observations of dynamic targets directly to actions of robotic manipulation. GA-DDPG~\cite{c12} and the teacher-student framework designed by Christen~\textit{et al.}~\cite{c13} demonstrate the feasibility of using 3D point clouds for policy learning, though they often rely on extra instance segmentation. 
Alternatively, some methods integrate grasp detection by encoding graspable points into the policy. For instance, GAMMA~\cite{c14} utilizes 9D features for online fusion, while EARL~\cite{c15} and GAP-RL~\cite{c16} leverage key points or stochastic Gaussian points to guide tracking and grasping. Generally, these approaches rely on high-speed inference to achieve passive-responsive tracking of moving targets.

However, instance segmentation models impose significant computational overhead, making them difficult to deploy on power-limited satellite platforms. Methods relying solely on tracking-and-approaching face challenges due to high-dynamic requirements and the need for extreme inference efficiency. In this letter, by extracting motion information through dual-frame correlation, our system enables target trend prediction and generates proactive manipulating actions without high computational costs and excessive hardware reliance.

\subsection{Imitation learning in robotic manipulation}
Imitation learning is a data-driven approach to robotic manipulation that learns policies directly from expert demonstrations~\cite{c18}. By eliminating the need for precise environmental models or accurate reward function engineering, it offers significant advantages in complex and unstructured scenarios~\cite{c19}.

Imitation learning encompasses various technical approaches, including Behavior Cloning~(BC)~\cite{c20}, Inverse Reinforcement Learning~(IRL)~\cite{c21}, Generative Adversarial Imitation Learning~(GAIL)~\cite{c22}, and Diffusion Policy~(DP)~\cite{c23}. Among these, Action Chunking with Transformers (ACT)~\cite{c29}, as a state-of-the-art method, BC and DP have emerged as focal points of current research due to their exceptional performance in robotic manipulation tasks. Although DP offers superior precision and generalization, its lengthy inference time makes it difficult to meet the demands of high-dynamic scenarios. 

This letter adopts ACT as a foundation and integrates an inter-frame correlation network to design an autonomous manipulation approach. 
The main contributions of this work are as follows. First, we propose a novel imitation learning architecture that captures spatio-temporal dependencies from continuous visual streams, effectively addressing the stability issues inherent in single-frame policies. Second, through extensive physical experiments, we validate the effectiveness of the proposed method, demonstrating significant improvements in both manipulation success rates and trajectory smoothness compared to baseline models. Finally, we developed a ground-based experimental platform capable of simulating 2D microgravity free-floating motion and single-axis rotation. Based on the platform, we conduct an in-depth analysis of the energy dissipation and dynamic processes during the manipulation of moving targets, providing a reliable physical basis for verifying autonomous manipulation strategies in space-like environments.

\begin{figure*}[h]
    \centering
    \includegraphics[width=\textwidth]{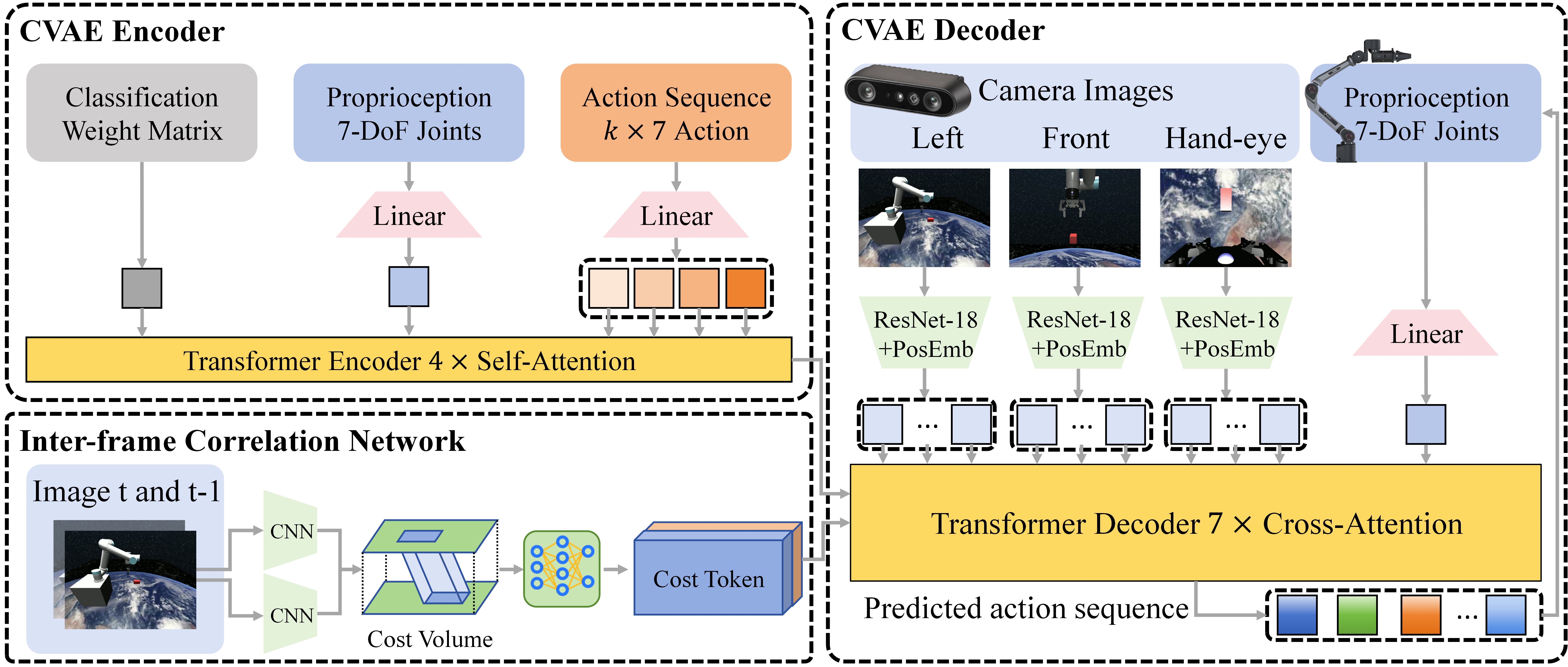}
    \caption{Overview of the proposed network architecture. The top-left portion displays the CVAE encoder, which predicts the style latent variable $z$ representing task characteristics. The bottom-left portion integrates an inter-frame correlation network to extract key motion tokens by comparing consecutive visual frames. The right portion shows the CVAE decoder, which maps the current multi-modal perception data and latent variables into the final robotic action sequences.}
    \label{fig:SystemStructure}
\end{figure*}

\section{Method}
To achieve precise manipulation of non-cooperative targets in complex and high-dynamic orbital environments, this chapter proposes an imitation learning architecture that integrates an Inter-frame correlation network. As illustrated in Fig.~\ref{fig:SystemStructure}, the system first extracts feature maps from two consecutive frames of a specific camera. An inter-frame correlation layer then computes the relationship between these maps and transforms the resulting spatio-temporal cues into high-dimensional tokens. These tokens, along with environmental representations, are fed into a downstream Transformer action prediction network, enabling the model to perceive motion trends and generate continuous joint control sequences~\cite{c24}. This design bypasses complex dynamic parameter estimation and significantly enhances the system's proactive interception capabilities in unstructured scenarios.

\subsection{Inter-frame Correlation Network}\label{sec:correlation}
\begin{figure*}
    \centering
    \includegraphics[width=\textwidth]{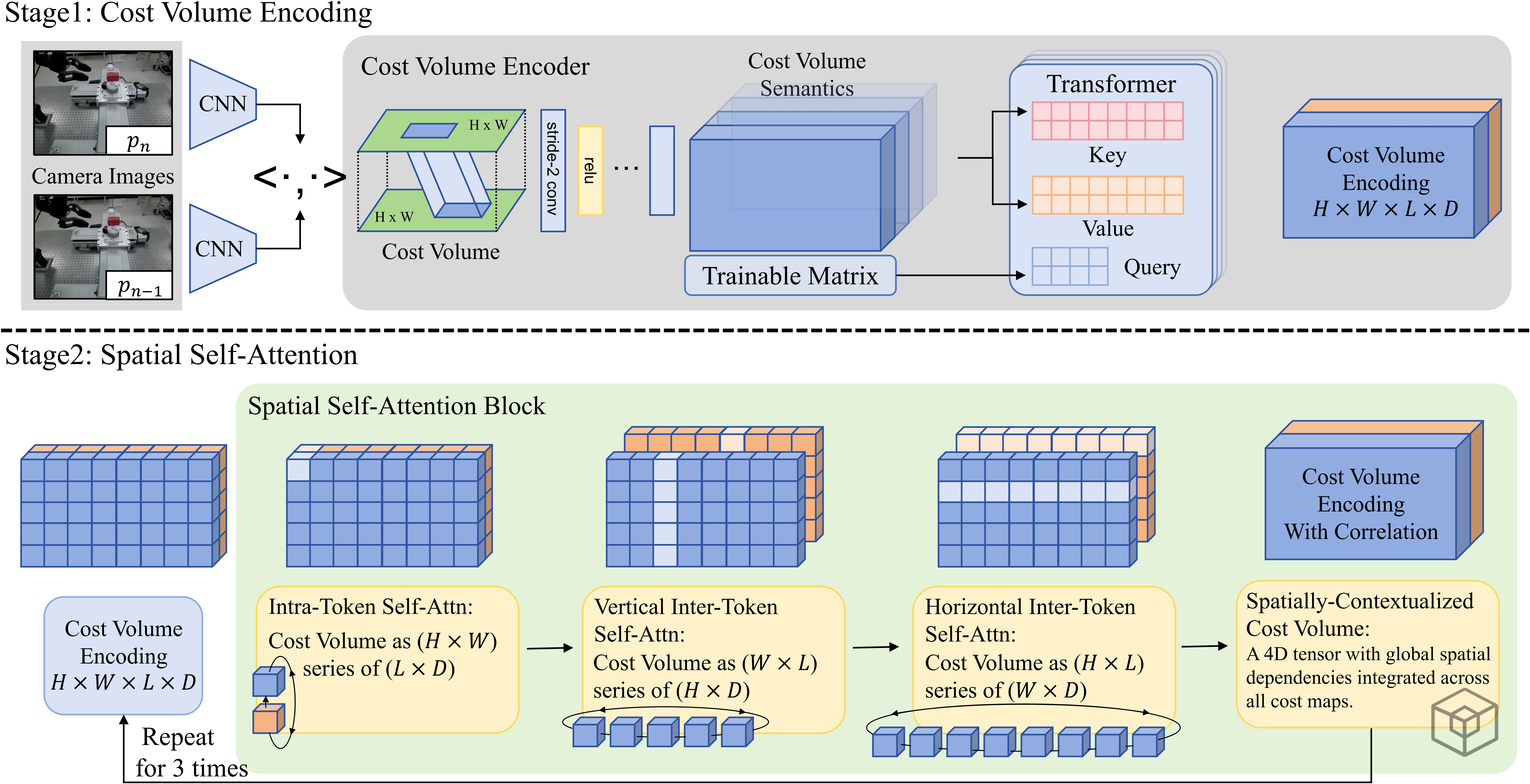}
    \caption{Schematic of the Inter-frame Correlation Network. The network processes spatio-temporal correlation information through two primary stages: the first stage involves cost volume calculation, encoding, and compression, transforming consecutive image features into cost volume semantic representations. The second stage implements a Spatial Self-Attention mechanism, utilizing intra-token, vertical, and horizontal self-attention blocks to enhance the network's ability to capture global spatial dependencies within the cost maps.}
    \label{fig:correlation}
\end{figure*}

We employ a 4D cost volume to model the relationships between two images, which is a widely used approach for measuring the similarity between various regions across image pairs~\cite{c25}. In this approach, when region A in Image 1 is found to be highly similar to region B in Image 2, it suggests that the object originally located at region A has moved to region B in the subsequent frame~\cite{c10,c30}. The process mentioned above illustrates how the cost volume serves as an effective tool for representing motion information within a scene, and the main structure of the inter-frame correlation network is shown in Fig.~\ref{fig:correlation}.

The structure of the inter-frame correlation network consists of two main stages.
In the first stage, a ResNet-18~\cite{c26} is employed as the backbone to extract deep features from the current image $I_t \in \mathbb{R}^{H' \times W' \times C}$ and the previous image $I_{t-1} \in \mathbb{R}^{H' \times W' \times C}$. By applying a $32\times$ downsampling factor, the backbone transforms these input images into high-level feature representations, ultimately yielding two feature tensors $F_t\in \mathbb{R}^{H \times W \times D}$ and $F_{t-1} \in \mathbb{R}^{H \times W \times D}$, where $H$, $W$, and $D$ denote the height, width, and channel depth of the feature maps, respectively.

In this method, the image resolution is set to $H'=480$ and $W'=640$, while $C$ represents the number of channels, the typical value being 3. The ResNet-18 backbone performs feature extraction with a $32\times$ downsampling factor, resulting in feature tensors $F_t, F_{t-1} \in \mathbb{R}^{15 \times 20 \times 512}$. Using these features, a 4D cost volume $\mathcal{C}$ is constructed by computing the all-pairs correlation between the two frames. This $15 \times 20 \times 15 \times 20$ volume exhaustively captures the matching confidence across all spatial regions and serves as the fundamental representation for motion information extraction. Each entry of the cost volume is computed via the dot product of feature vectors at their respective spatial locations, as shown in Eq.~\eqref{eq:cost_volume}:
\begin{equation}
\label{eq:cost_volume}
\mathbf{\mathcal{C}}(i, j, k, l) = \mathbf{F}_t(i, j)^\top \mathbf{F}_{t-1}(k, l)
\end{equation}

In the second stage, the 4D cost volume $\mathbf{\mathcal{C}}$ is reshaped into a set of $15 \times 20$ correlation matrices, each with a spatial dimension of $15 \times 20$. Each matrix represents the similarity distribution between a specific location in the current frame and the entire search space in the reference frame. To extract high-level representations from these distributions, a multi-layer Convolutional Neural Network (CNN) is employed to generate cost volume embeddings, the details of which are detailed in Table~\ref{tab:cnn_arch} and Eq.~\ref{eq:CNN}.
\begin{equation}
\label{eq:CNN}
\mathbf{\mathcal{S}}_{i,j} = \text{CNN}(\mathbf{\mathcal{C}}_{i,j})
\end{equation}

\begin{table}[h]
\centering
\caption{Architecture of the CNN for cost volume Embedding.}
\label{tab:cnn_arch}
\begin{tabular}{lcccc}
\toprule
Layer & Input Size & Kernel & Stride & Channel \\ 
\midrule
Conv1 + ReLU & $15 \times 20$ & $6 \times 6$ & 2 & $d_{\text{embed}}/4$ \\
Conv2 + ReLU & $7 \times 10$ & $6 \times 6$ & 2 & $d_{\text{embed}}/2$ \\
Conv3        & $3 \times 5$  & $6 \times 6$ & 2 & $d_{\text{embed}}$ \\
\bottomrule
\end{tabular}
\end{table}

Through a series of strided convolutions, the CNN reduces the spatial resolution of each correlation matrix from $15 \times 20$ to $2 \times 3$, which is then flattened to form a semantic feature vector of length $6$. This results in a semantic representation $ \mathbf{\mathcal{S}}\in \mathbb{R}^{15 \times 20 \times 6 \times d_{\text{embed}}}$. 

Given the inherent information redundancy in this representation, a Transformer cross-attention mechanism for feature compression is introduced to explore the interactions between different spatial locations. Each $\mathbf{\mathcal{S}}_{i,j} \in \mathbb{R}^{6 \times d_{\text{embed}}}$ is treated as Key and Value, while two $d_{\text{embed}}$-dimensional trainable latent queries serve as the Query. This mechanism refines and compresses the information into a compact cost volume encoding of size $15 \times 20 \times 2 \times d_{\text{embed}}$, providing a more efficient input for downstream motion estimation tasks.

Subsequently, a spatial self-attention mechanism is introduced to capture the intra-token relationships and spatial dependencies within the cost volume tokens, where the $15 \times 20 \times 2 \times 512$ cost volume encoding is first interpreted as $300$ local vectors of size $2 \times 512$, performing self-attention within each vector. To further capture global spatial contexts, the encoding is then rearranged into two separate sequences of size $15 \times 20 \times 512$. For each sequence, inspired by the axial decomposition strategy in Swin Transformer~\cite{c27,c31}, we treat it as $15$ horizontal sequences of length $20$ and $20$ vertical sequences of length $15$, performing self-attention along each axis sequentially. These operations collectively constitute one spatial self-attention block, which is repeated three times. This hierarchical and decomposed attention mechanism enables the model to efficiently integrate local correlations and global topological information with significantly reduced computational complexity. Finally, the cost volume encoding is flattened into $600$ cost volume tokens, each with a feature dimension of $512$, and fed into the policy network.

\subsection{Policy Training}
The backbone of the proposed model is a CVAE-Transformer architecture, and during training, the CLS token, proprioception, and the target action sequence are embedded into a common dimension $d_m$ via linear layers and concatenated to form the non-visual observation $\tilde{O}_t$. This input is processed by a $4$-layer Transformer encoder $q(z|\tilde{O}_t)$ to yield the latent variable $z$. In the VAE context, $z$ represents the latent distribution, while in the imitation learning setting, it serves as a style variable characterizing the specific action style.

Regarding the decoder, three camera views, containing two global and one hand-eye, are processed by independent pre-trained ResNet-18 backbones to extract features, yielding three $15 \times 20 \times 512$ feature tensors. These features are flattened into a sequence of $900$ visual tokens. These are concatenated with the proprioception, the style variable $z$, projected to dimension $d_m$ via linear layers, and $600$ cost volume tokens, which characterize scene motion, to form the environment observation $O_t$, consisting of $1502$ tokens in total. The decoder is defined as $\pi(a_{t:t+k}|O_t)$, comprising a $4$-layer Transformer encoder and a $7$-layer Transformer decoder. The transformer decoder's Query uses fixed positional embeddings, while its Key and Value are derived from the transformer encoder's output. The model predicts a sequence of actions for the next $k$ steps. As formulated in Eq.~\eqref{eq:total_loss}, the network is trained using a weighted combination of $L_1$ loss and KL divergence, optimized via the AdamW optimizer.
\begin{equation}
\label{eq:total_loss}
\mathcal{L} = \frac{1}{7k}\sum_{i=t}^{t+k} \| \hat{a}_i - a_i \|_1 + \beta D_{KL}(q(z|\tilde{O}_t) \parallel \mathcal{N}(0, I))
\end{equation}

\subsection{Policy Evaluation}
Once training is completed, the CVAE encoder is removed, and the style variable $z$ is set to a zero matrix. The remaining CVAE decoder serves as the policy network $\pi(a_{t:t+k}|O_t)$. During inference, the environment observation $O_t$ is formed by concatenating $z$, proprioception, image features from the three camera views, and the cost volume tokens derived from the image stream of one specific camera via the temporal correlation described in Section~\ref{sec:correlation}. The policy then predicts a sequence of $k$ future actions in a single forward pass. To ensure smooth motion, we employ temporal aggregation, which calculates a weighted sum of overlapping action sequences using an exponential weighting scheme $w_i = \exp(-k \times i)$, while $i=0$ represents the weight $w_0$ for the most recently predicted action. In this manner, the network not only produces smooth action sequences but also demonstrates enhanced robustness in dynamic environments by prioritizing the most recent predictions.

\section{Experiment}
This chapter provides a comprehensive evaluation of the proposed strategy through seven core experiments. Beyond the baseline task, we simulate complex orbital conditions via low-light, random camera occlusion, and random target occlusion tests. To assess dynamic adaptability, we designed a target maneuver test to observe responses to sudden motion changes. Furthermore, a Monte Carlo experiment was conducted to quantify the model's sensitivity to initial positions, while a grayscale vision test evaluated the system's reliance on geometric features and temporal correlations in the absence of color information. Together, these experiments analyze the model's smoothness, proactivity, and robustness under diverse sensing interferences and dynamic scenarios.

\subsection{Experimental Platform}
\begin{figure}[htbp]
    \centering
    \includegraphics[width=\linewidth]{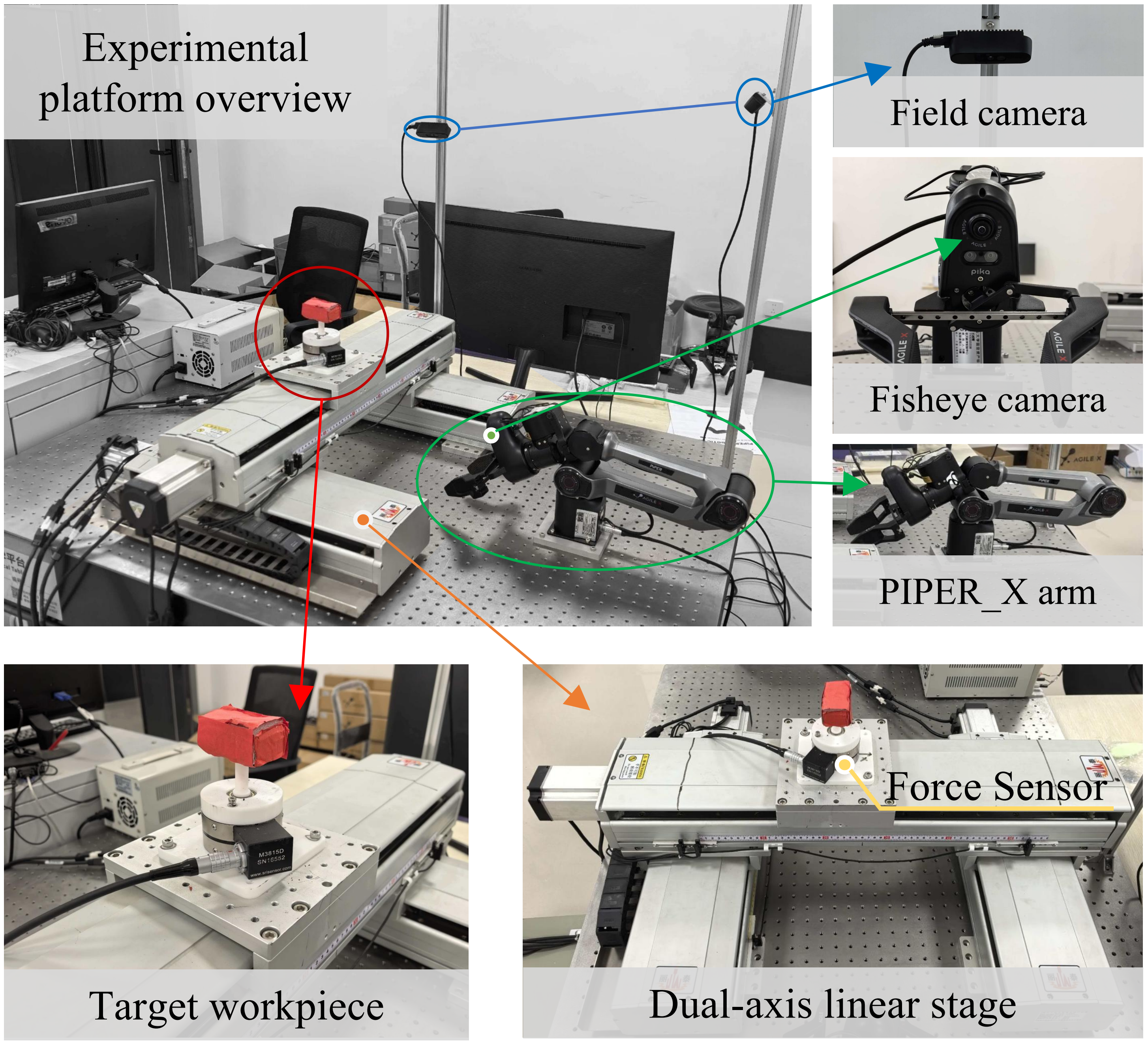} 
    \caption{Overview of the ground-based experimental platform.}
    \label{fig:ground_platform}
\end{figure}

To evaluate the proposed algorithm's capability in perceiving and manipulating dynamic targets in a ground-based environment, an experimental platform was developed. The platform primarily consists of three components: the execution unit, the motion simulation unit, and a multi-view perception system, as shown in Fig.~\ref{fig:ground_platform}. The core execution unit is a $6$-DOF PIPER\_X lightweight robotic arm. To replicate the microgravity free-floating characteristics of orbital targets, the workpiece is mounted on a dual-axis linear stage via a 6-axis force sensor and a low-friction bearing, allowing for translation in the $X$-$Y$ plane and free rotation. Based on an active compliance strategy, the system captures interaction forces $\mathbf{F}$ in real-time upon contact. It then calculates velocity increments using the momentum theorem, $\mathbf{F} \Delta t = m \Delta \mathbf{v}$, and drives the stage to execute compensatory movements, effectively simulating microgravity behaviors such as inertial gliding and collision rebound. 

\begin{figure}[htpb]
    \centering
    \includegraphics[width = \linewidth]{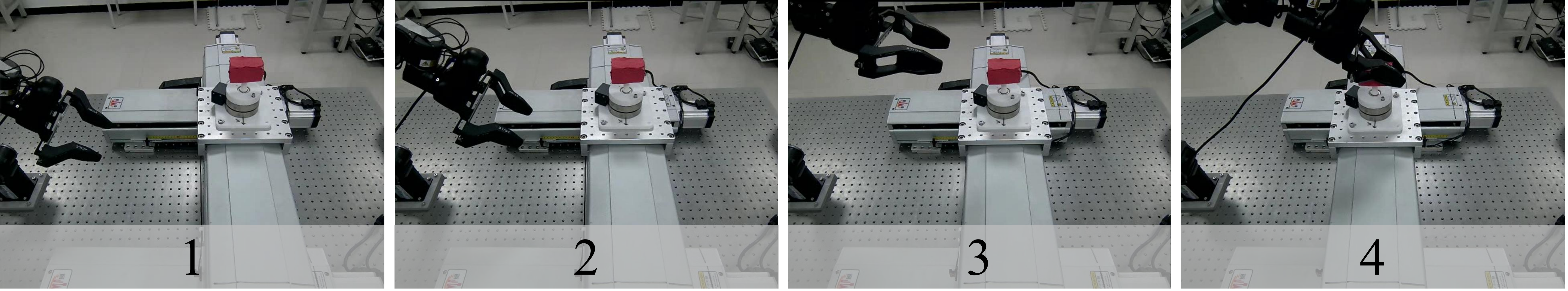}
    \caption{Workflow of the experiment.}
    \label{fig:workflow}
\end{figure}

At the beginning of each experiment, the target moves toward the robotic arm with a random velocity, which is between $0.6\,\text{cm/s}$ and $1.2\,\text{cm/s}$ with a random direction. The robotic arm must autonomously grasp the moving object and bring it to a stop without any prior knowledge, as shown in Fig.~\ref{fig:workflow}. To evaluate how robotic arm oscillations affect the attitude stability of the spacecraft, we set up a space robotic manipulation scenario and used a simplified dynamic model to analyze the relationship between joint vibration and spacecraft disturbance. In this study, we use Mean Absolute Second Difference (MASD) to measure the oscillation level and set a maximum allowable MASD threshold for each joint. A task is judged as successful only if the robotic arm successfully grasps the target, stops it, and keeps the MASD within the threshold throughout the process.

\begin{figure}[htbp]
    \centering
    \includegraphics[width=\linewidth]{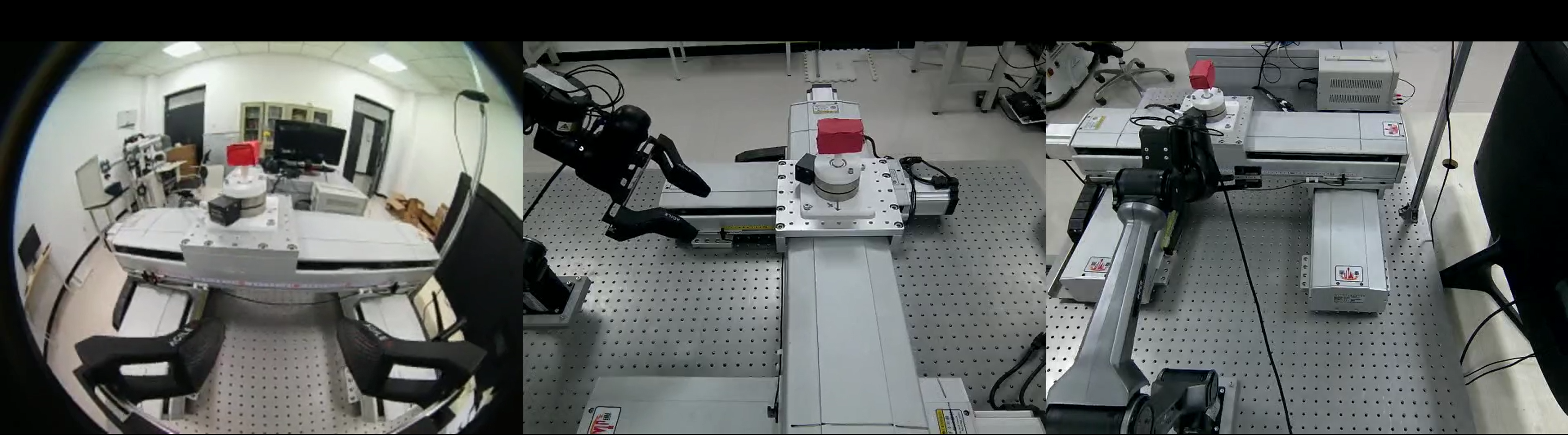}
    \caption{This figure illustrates the multi-view visual feedback during manipulation. From left to right are the hand-eye camera 
    view and two global field camera views.}
    \label{fig:CameraView}
\end{figure}

For environmental perception, the setup employs a multi-view visual feedback system, including two fixed-field cameras and a hand-eye fisheye camera, as illustrated in Fig.~\ref{fig:CameraView}. The hand-eye camera provides a close-up view of the target and the end-effector, while the two global cameras offer broader perspectives of the workspace. This multi-view configuration allows the system to capture comprehensive visual information, enabling robust perception and accurate manipulation of dynamic targets under various conditions.

\subsection{Experiment Results}
To measure how robotic arm motion affects the base attitude, we built a dynamic model of capture, with the simulation conditions as follows. It is assumed that this model consists of a $300\,\text{kg}$ micro-satellite and a PIPER\_X robotic arm. To keep the satellite's attitude stable, the system uses reaction wheels with a maximum torque of $0.1$ Nm. Based on the law of conservation of momentum, the reaction torque $\tau_{\text{react}}$ created by joint motion must be smaller than the maximum torque of the wheels to avoid attitude instability. Based on the angular momentum equations of the system, the reaction torque of joint $j$ can be written as $\tau_{\text{react},j} = I_{eq,j}(\theta) \cdot \alpha_{j} $, where $I_{eq,j}$ is the equivalent moment of inertia of joint $j$ in a fully extended state, and $\alpha_{j}$ is the joint angular acceleration. 

By calculating the inertia of each joint under the worst conditions, we obtained the maximum allowable acceleration limits. In this study, we use these physical constraints directly as the MASD thresholds for task success. The base joint (J1) and the shoulder joint (J2) are the most restricted, with maximum allowable accelerations of $0.178\,\text{rad/s}^2$ and $0.255\,\text{rad/s}^2$, respectively. In contrast, starting from the elbow joint (J3), no additional safety limits are set because the disturbance is far below the wheel compensation limit due to the significant drop in equivalent inertia. During experiments, if the oscillations of J1 or J2 exceed these limits, the task is judged as a failure because the disturbance goes beyond the compensation range of the reaction wheels~\cite{c34}.

\begin{figure}[htpb]
    \centering
    \includegraphics[width=\linewidth]{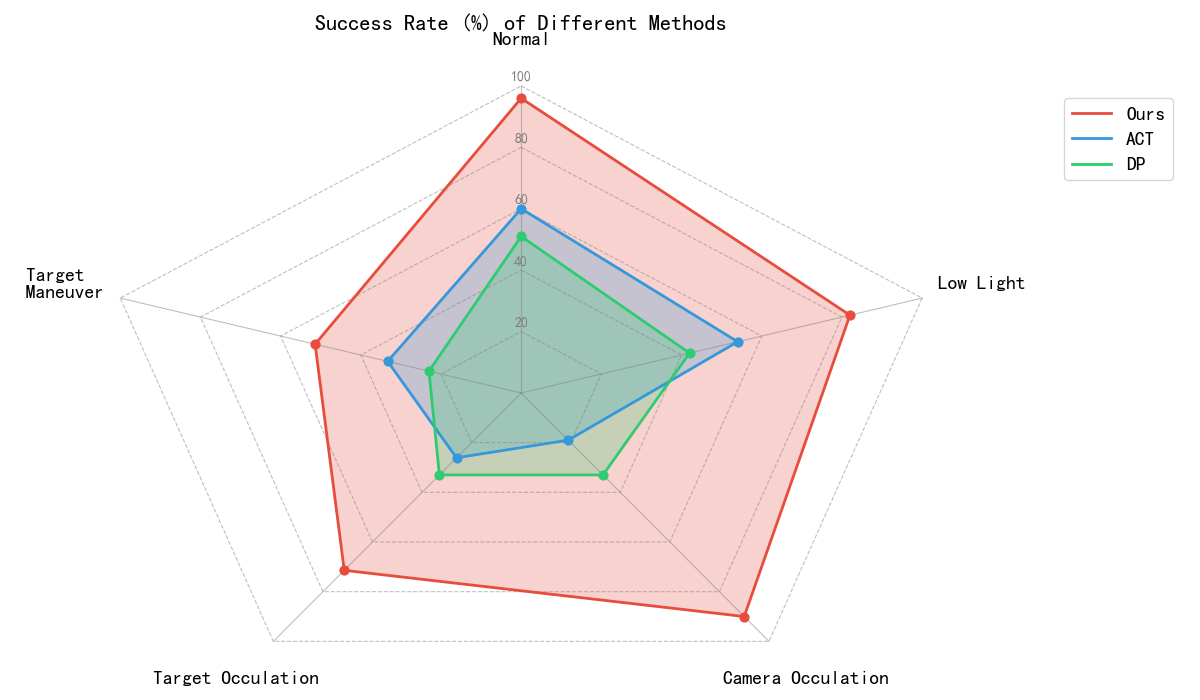}
    \caption{Radar chart comparing the performance of our method, the vanilla ACT, and DP across five scenarios.}
    \label{fig:RadarChart}
\end{figure}

We evaluated our proposed method against several baselines across five scenarios: Standard, Low-light, Camera Occlusion, Target Occlusion, and Target Maneuver. As shown in the experimental results, our method achieved success rates of 96\%, 82\%, 90\%, 71.4\%, and 51.4\%, respectively. In contrast, the vanilla ACT exhibited a noticeable decline in performance across these scenarios, particularly under challenging conditions such as occlusions. When the safety constraints for manipulation smoothness were applied, its success rates were recorded at 60\%, 54\%, 19\%, 26.1\%, and 33.3\%, respectively. A comprehensive performance comparison across these dimensions is illustrated in the radar chart in Fig.~\ref{fig:RadarChart}.

\begin{figure}
    \centering
    \includegraphics[width=\linewidth]{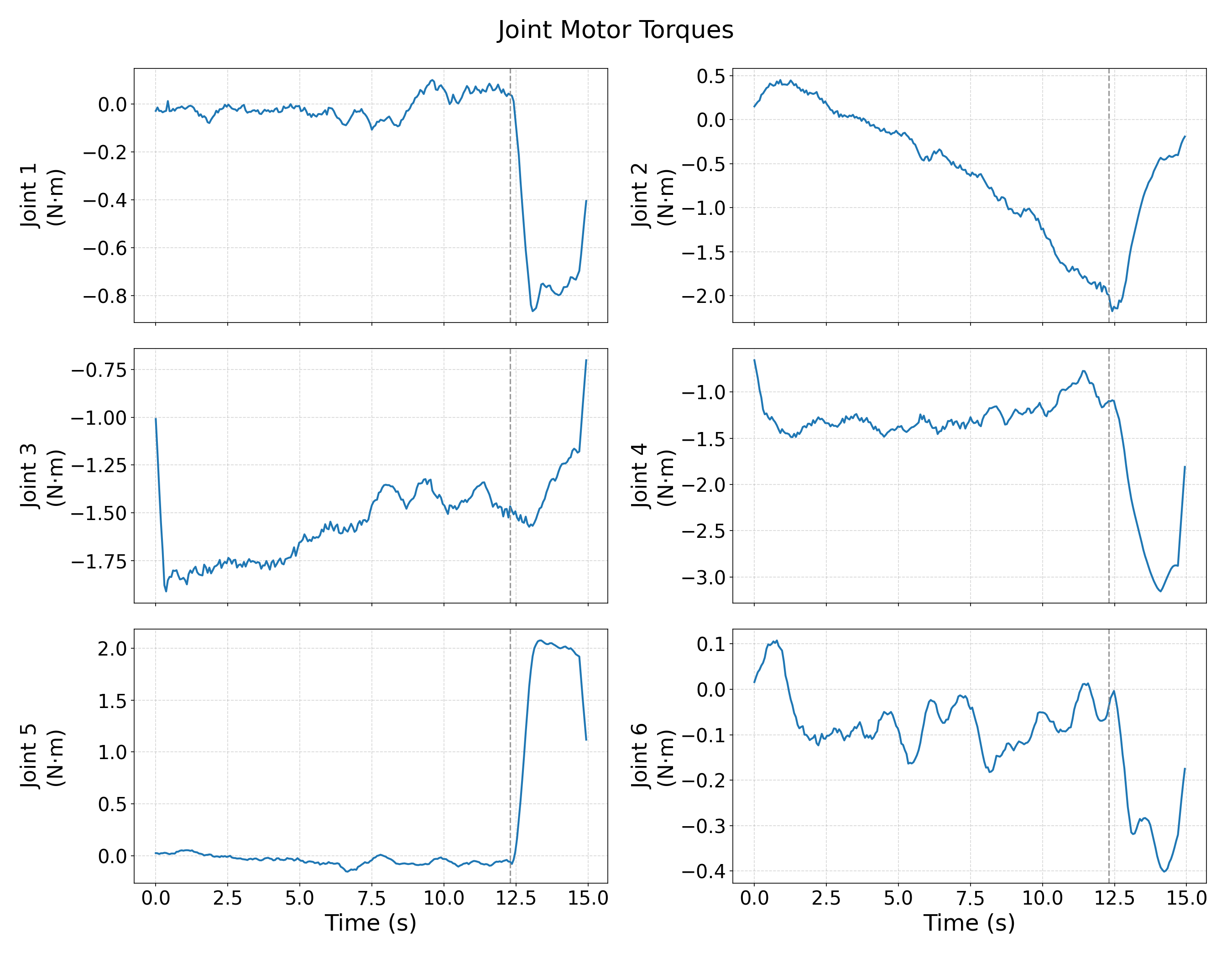}
    \caption{Motor torque profiles during a successful manipulation task. The gray dashed line indicates the moment of end-effector closure. It is observed that the motor torque increases immediately at this instant to provide the necessary constraint to bring the target to a halt.}
    \label{Torque}
\end{figure}

Upon successful capture, the robotic arm increases torque to establish a rigid constraint, bringing the target to a halt, and the corresponding motor torque profiles are shown in Fig.~\ref{Torque}. 
\begin{figure}[htpb]
    \centering
    \includegraphics[width=\linewidth]{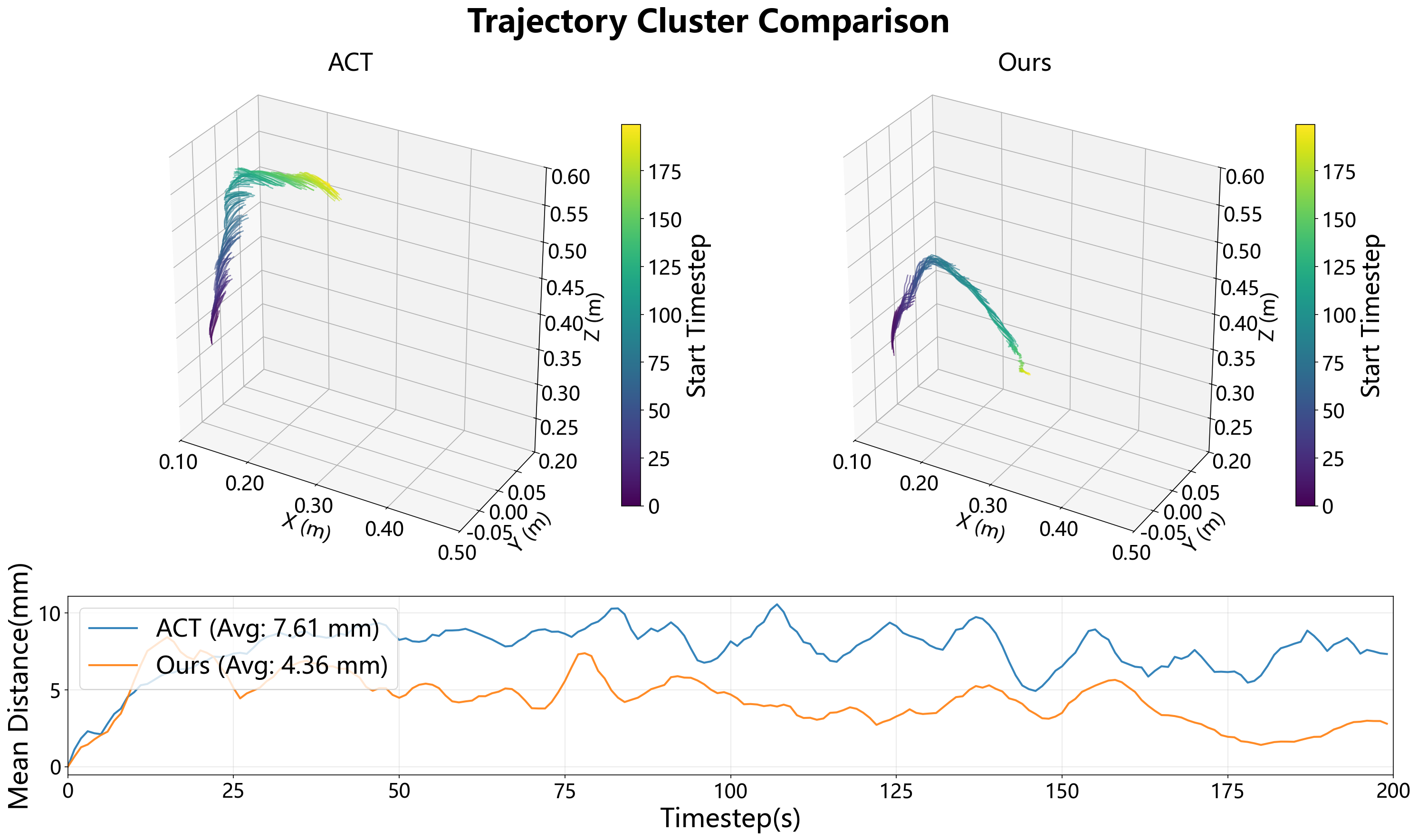}
    \caption{Comparison of trajectory clusters and divergence between the vanilla ACT (left) and our method (right). 
    The bottom plot displays the real-time ``average distance'' curves, with the legend indicating the time-averaged mean distance for each method to quantify overall cluster divergence.}
    \label{TrajectoryCluster}
\end{figure}

To analyze how capturing motion features reduces oscillations, we visualized the trajectory clusters, the set of all predicted action horizons generated at each time step, for both our method and the vanilla ACT shown in Fig.~\ref{TrajectoryCluster}. We quantified cluster divergence by calculating the time-averaged distance of all predicted points from the mean trajectory across the task duration. As illustrated, our method achieves significantly higher trajectory concentration, with predicted paths tightly clustered around the mean. We also evaluated the DP method. However, due to its slow inference speed, DP cannot utilize real-time inference or temporal aggregation. This leads to extreme policy oscillations. Consequently, we calculated a ``relaxed'' success rate for DP that ignores the oscillation criteria, resulting in scores of 51\%, 42\%, 33\%, 33\%, and 23\%.

\begin{figure*}[t]
\centering
\captionsetup{type=table}
\caption{Quantitative comparison of policy oscillations.}
\label{tab:combined_results}

\begin{tabularx}{\textwidth}{@{} l *{10}{C} @{}}
\toprule
\multirow{2}{*}{\textbf{Method}} & \multicolumn{2}{c}{\textbf{Normal}} & \multicolumn{2}{c}{\textbf{Camera Occlusion}} & \multicolumn{2}{c}{\textbf{Target Occlusion}} & \multicolumn{2}{c}{\textbf{Target Maneuver}} & \multicolumn{2}{c}{\textbf{Low-light}} \\
\cmidrule(lr){2-3} \cmidrule(lr){4-5} \cmidrule(lr){6-7} \cmidrule(lr){8-9} \cmidrule(lr){10-11}
& MASD($\downarrow$)   & Jerk($\downarrow$)   & MASD($\downarrow$)   & Jerk($\downarrow$)   & MASD($\downarrow$)   & Jerk($\downarrow$)   & MASD($\downarrow$)   & Jerk($\downarrow$)   & MASD($\downarrow$)   & Jerk($\downarrow$)   \\
\midrule
DP & 13.73 & 37.99 & 14.22 & 36.86 & 16.07 & 37.98 & 14.79 & 36.21 & 15.88 & 38.04 \\
Vanilla ACT & 0.231 & 8.87 & 0.236 & 9.26 & 0.269 & 11.03 & 0.241 & 8.26 & 0.259 & 10.42 \\
\textbf{Ours} & \textbf{0.167} & \textbf{5.71} & \textbf{0.199} & \textbf{6.48} & \textbf{0.181} & \textbf{6.33} & \textbf{0.167} & \textbf{5.86} & \textbf{0.204} & \textbf{7.64} \\
\bottomrule
\end{tabularx}
\end{figure*}

Given experimental results, our method satisfied the manipulation smoothness requirements, demonstrating the reliability of our method in fundamental tasks. Table~\ref{tab:combined_results} provides the MASD and RMS Jerk of the actions generated by different methods in typical scenarios, where these metrics are calculated as the average across all joints.

In the previous test, the initial position of the target was fixed, and its velocity was set to be higher than $1.35\, \text{cm/s}$. To evaluate the robustness of our method in a challenging environment, we performed a Monte Carlo experiment, which requires the robotic arm to perform autonomous manipulation starting from various initial positions near its default starting point and recorded the success rates for each position.
\begin{figure}[htpb]
    \centering
    \includegraphics[width=\linewidth]{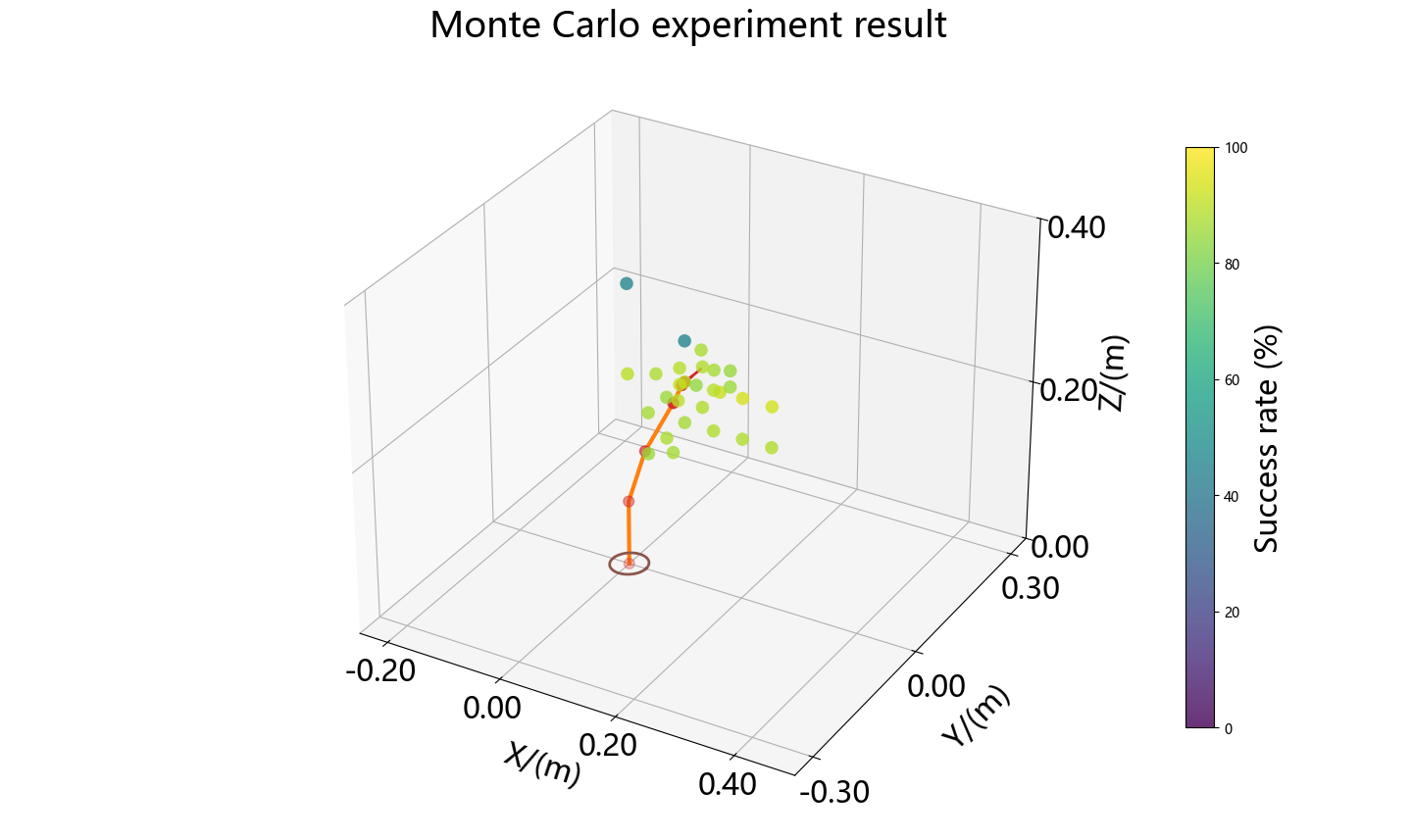}
    \caption{Monte Carlo experiment results showing the success rates of the robotic arm from various initial positions.}
    \label{MonteCarlo}
\end{figure}

As shown in Fig.~\ref{MonteCarlo}, the system can manipulate the target with a very high success rate across almost the entire range. Only at a few positions too far from the target was the robotic arm unable to respond in time due  to physical reach limits. These results quantify that the model has low sensitivity to changes in initial positions and shows  excellent stability.

Finally, we designed and performed a grayscale testing experiment. This experiment aims to simulate scenarios with significant visual information loss, such as extreme lighting conditions in space shadows, by converting real-time images into grayscale before feeding them into the network trained exclusively on color data.
\begin{figure}[htpb]
    \centering
    \includegraphics[width=\linewidth]{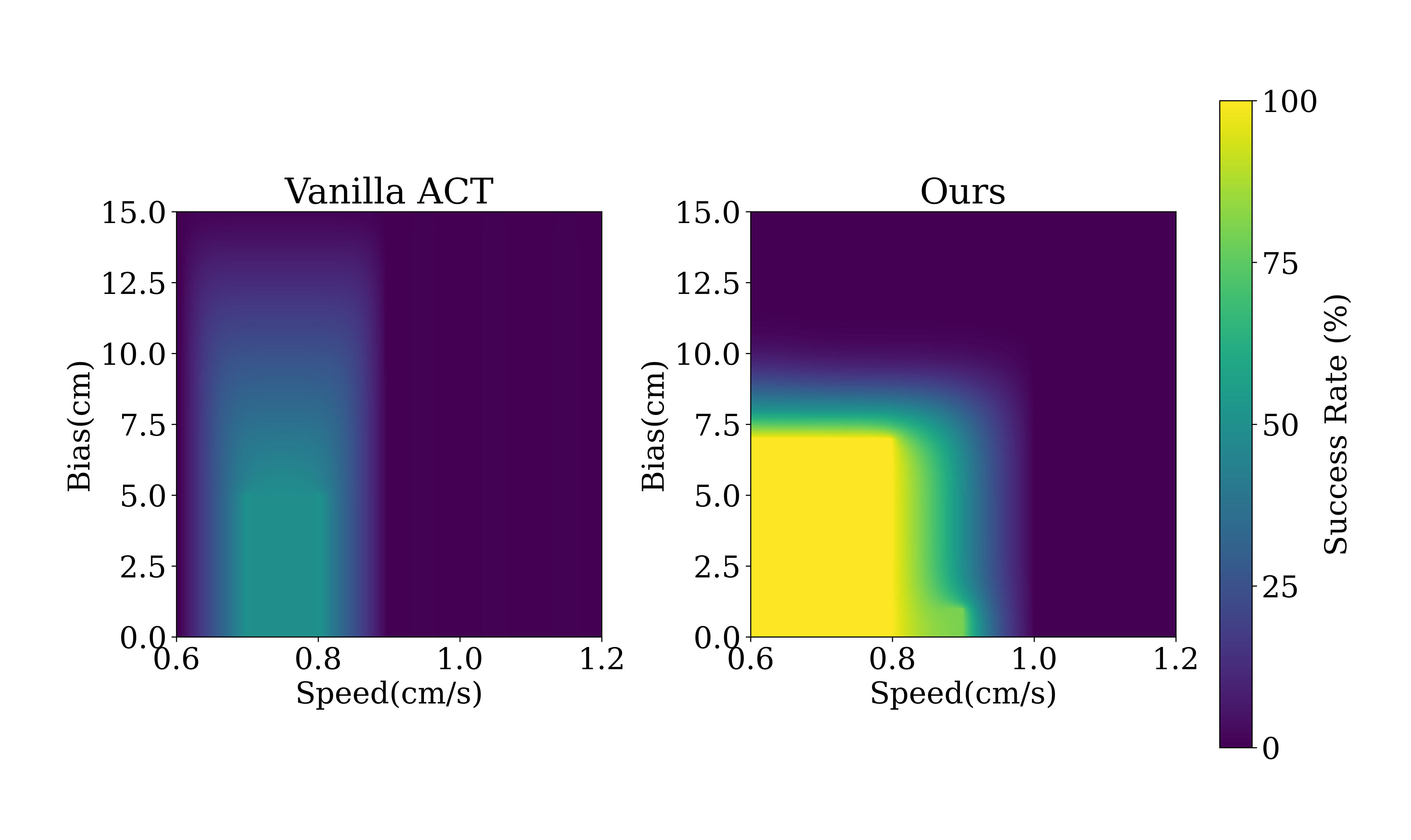}
    \caption{Heat map illustrating the success rates of the robotic manipulation across different initial target positions in the grayscale testing.}
    \label{HeatMap}
\end{figure}

Experimental results show that in this challenging scenario, both our method and the vanilla ACT experienced a significant decrease in success rate, specifically manifesting as a shrinkage of the effective manipulation workspace relative to the target's initial position. To intuitively compare the performance of the two methods, we plotted heatmaps illustrating the relationship between the target's initial position offset, which is relative to the robotic arm's axis, velocity, and the manipulation success rate, as shown in Fig.~\ref{HeatMap}. It can be clearly observed from the heatmaps that our method maintains a significantly larger manipulation workspace than ACT under these conditions.

\section{CONCLUSIONS}
This paper proposes a robust imitation learning method for autonomous grasping in dynamic space environments by integrating an Inter-frame Correlation Network. By explicitly modeling temporal motion features, the proposed framework effectively resolves the action ambiguity inherent in single-frame observations when encountering multi-modal distributions. Experimental results demonstrate that, compared to the vanilla ACT, our method significantly suppresses policy oscillations and enhances trajectory concentration and temporal consistency. Crucially, under stringent dynamic safety constraints, which are the acceleration limits of the base joint (J1), our method maintains high success rates across diverse scenarios, including low-light, occlusions, and target maneuvers. In contrast, baseline methods are susceptible to attitude instability of the spacecraft due to dynamic coupling. Furthermore, our method exhibits superior robustness under various disturbance conditions.
Specifically, in both the Monte Carlo and grayscale experiments, the system inputs were out-of-distribution relative to the training data. The performance of our method in these experiments involving distribution shifts demonstrates its excellent 
generalization capability.

In future work, we will further explore the potential of this high-performance architecture by extending its application to diverse spacecraft platforms, such as planetary surface robots. Additionally, as advancements in space microelectronics lead to increased computational power for on-board satellite devices, we plan to incorporate Large Language Models (LLMs) to evolve the framework into a Vision-Language-Action (VLA) model, striving to enable space robots to perform intelligent, long-horizon, and multi-task manipulation.

\addtolength{\textheight}{-5cm}

\small
\bibliography{reflist.bib}

\end{document}